%% file: main.tex
\documentclass{article}

\usepackage[preprint,nonatbib]{neurips_2019}

\usepackage[utf8]{inputenc} 
\usepackage[T1]{fontenc}    
\usepackage{hyperref}       
\usepackage{url}            
\usepackage{booktabs}       
\usepackage{amsfonts}       
\usepackage{nicefrac}       
\usepackage{microtype}      

\usepackage{graphicx}
\usepackage{amsmath}
\usepackage{amssymb}
\usepackage{subfig}
\usepackage{mathtools}
\mathtoolsset{showonlyrefs}
\DeclareMathOperator*{\argmax}{arg\,max}

\newcommand{\rd}{\mathrm{d}}
\newtheorem{lemma}{\noindent Lemma}
\newtheorem{theorem}{\noindent Theorem}
\newtheorem{definition}{\noindent Definition}
\newtheorem{example}{\noindent Example}
\newtheorem{proposition}{\noindent Proposition}
\usepackage{todonotes}
\usepackage{multirow}
\usepackage{algorithm}
\usepackage{algorithmic}

\title{Wasserstein-Fisher-Rao Document Distance}

\author{%
  	Zihao Wang \\
  	Department of Computer Science and Technology\\ Tsinghua University\\
    \texttt{wzh17@mails.tsinghua.edu.cn} \\
    \And
  	Datong Zhou \\
	Department of Mathematical Science\\
	Tsinghua University\\
	\texttt{zdt14@mails.tsinghua.edu.cn} \\
	\And
	Yong Zhang\\
	Department of Computer Science and Technology \\ Tsinghua University\\
	\texttt{zhangy05@tsinghua.edu.cn} \\
	\And
	Hao Wu \\
	Department of Mathematical Science\\
	Tsinghua University\\
	\texttt{hwu@tsinghua.edu.cn} \\
	\And
	Chenglong Bao \\
	Yau Mathematical Sciences Center\\
	Tsinghua University, China\\
	clbao@math.tsinghua.edu.cn
}

\begin{document}

\maketitle

\begin{abstract}

As a fundamental problem of natural language processing, it is important to measure the distance between different documents. Among the existing methods, the Word Mover’s Distance (WMD) has shown remarkable success in document semantic matching for its clear physical insight as a parameter-free model. However, WMD is essentially based on the classical Wasserstein metric, thus it often fails to robustly represent the semantic similarity between texts of different lengths. In this paper, we apply the newly developed Wasserstein-Fisher-Rao (WFR) metric from unbalanced optimal transport theory to measure the distance between different documents. The proposed WFR document distance maintains the great interpretability and simplicity as WMD. We demonstrate that the WFR document distance has significant advantages when comparing the texts of different lengths.
The varying length matching and KNN classification results on eight datasets have shown its clear improvement over WMD. Furthermore, WFR could also improve WMD under other frameworks.

\end{abstract}

\input{src/intro.tex}
\input{src/related.tex}
\input{src/method.tex}

\input{src/discuss.tex}

\input{src/conclusion.tex}
\newpage
\bibliography{main.bib}
\bibliographystyle{acm}

\input{src/appendix.tex}

\end{document}

%% file: src/intro.tex
\section{Introduction}

Measuring the similarity between documents plays an important role in natural language processing. Recently, Word Mover's Distance (WMD)~\cite{DBLP:conf/icml/KusnerSKW15}, as a metric in probability space, has clear interpretation, solid theoretical foundation and demonstrated great success in many applications, e.g.\ metric learning~\cite{DBLP:conf/nips/HuangGKSSW16}, document retrieval~\cite{DBLP:conf/emnlp/WuYXXBCRW18}, question answering~\cite{DBLP:conf/bionlp/BrokosMA16} and machine translation~\cite{DBLP:conf/aaai/Zhang0LSIH16}. 
More concretely, in the word embedding space,
WMD employs the Wasserstein metric on the space of normalized bag of words (nBOW) distribution of documents, i.e.\
given two documents $D_s = \{x^s_1, ..., x^s_m\} $ and $ D_t = \{x^t_1, ..., x^t_n\} $ with nBOW distributions $ f^s $ and $f^t$,  the WMD of $D_s$ and $D_t$ is 
\begin{equation}
\begin{aligned}
\mathrm{WMD}(D_s,D_t) = \min_{R\in\mathbb{R}^{m\times n}}\left\{\sum_{ij} C_{ij}R_{ij}\left|\sum_j R_{ij} = f^s_i, \sum_i R_{ij} = f^t_j\right.\right\}
\end{aligned}
\end{equation}
where $C_{ij}=\lVert x_i^t - x_j^s \rVert$ is the transport cost and $ x_{\cdot}^{\cdot} $ is the word vector.
With the help of optimal transport, WMD naturally bridges the document distance and the word similarity in the embedding space.
Moreover, it is worth noting	 that  optimal transport metric (Wasserstein metric) has shown many new insights in generative adversarial networks~\cite{DBLP:conf/icml/ArjovskyCB17}, domain adaptation~\cite{DBLP:conf/aaai/ShenQZY18}, representation learning~\cite{DBLP:conf/nips/MuzellecC18}, and etc.
\begin{figure}
	\centering
	\includegraphics[width=0.6\textwidth]{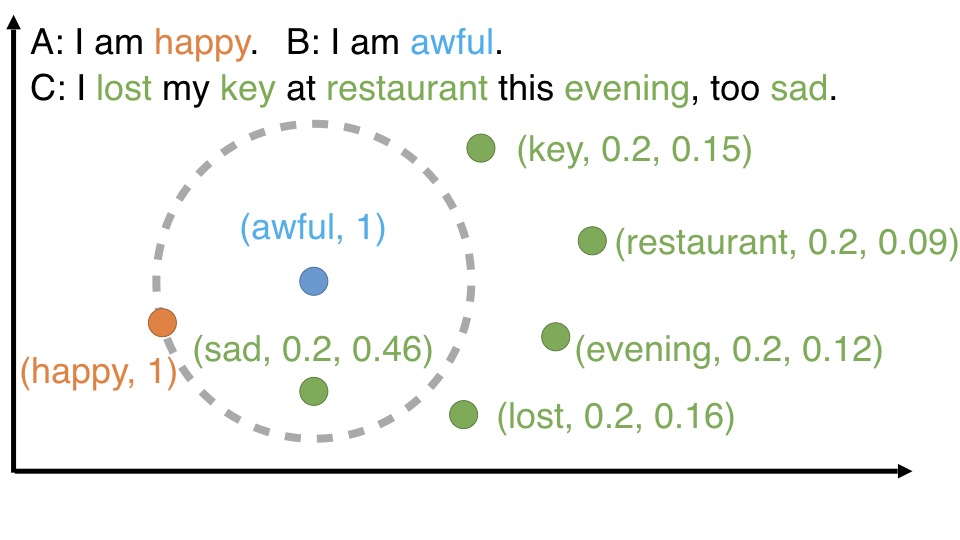}
	\vskip-2em
	\caption{Illustration of transport plans by WMD (Example \ref{Example:WMD}) and WFR Document Distance (Example \ref{Example:WFR}). ``(key, 0.2, 0.15)'' denotes the mass of the word ``key'' is 0.2 in WMD while 0.15 in WFR.}
	\label{fig-demo}
\end{figure}

\begin{table*}[t]
\caption{Transport plan by Example~\ref{Example:WMD} (WMD) and Example~\ref{Example:WFR} (WFR)}
\vskip-.5em
	\label{Table:example}
	\centering
	\small
\begin{tabular}{ccllllllll}
\toprule
\multicolumn{2}{c}{\multirow{4}{*}{word}} & \multicolumn{4}{c}{WMD} & \multicolumn{4}{c}{WFR} \\
\cline{3-10}
\multicolumn{2}{c}{} & \multicolumn{2}{c}{B} & \multicolumn{2}{c}{cost} & \multicolumn{2}{c}{B} & \multicolumn{2}{c}{cost}                           \\
\cline{3-10}
\multicolumn{2}{c}{} & \multicolumn{2}{c}{awful} & \multirow{2}{*}{amount} & \multirow{2}{*}{total} & \multicolumn{2}{c}{awful}                                  & \multirow{2}{*}{amount} & \multirow{2}{*}{total} \\
\cline{3-4} \cline{7-8}
\multicolumn{2}{c}{} & \multicolumn{1}{c}{indiv. cost} & \multicolumn{1}{c}{mass} & \multicolumn{2}{c}{} & \multicolumn{1}{c}{indiv. cost} & \multicolumn{1}{c}{mass} & \multicolumn{2}{c}{} \\
\hline
\multicolumn{1}{l}{A} & happy & 1.43 & 1 & 1.43 & 1.43 & 2.04                            & 1 & 2.04 & 2.04 \\
\hline
\multirow{5}{*}{C} & sad & 1.20 & 0.20 & 0.24 & \multirow{5}{*}{1.50} & 1.45 & 0.47 & 0.68 & \multirow{5}{*}{1.96} \\
 & lost & 1.49 & 0.20 & 0.30 & & 2.21 & 0.16 & 0.35 & \\
 & key & 1.50 & 0.20 & 0.30 & & 2.25 & 0.15 & 0.34 & \\
 & evening & 1.56 & 0.20 & 0.31 & & 2.42 & 0.12 & 0.30 & \\
 & restaurant & 1.75 & 0.20 & 0.35 & & 3.07 & 0.09 & 0.30 & \\
\bottomrule
\vspace{-3em}
\end{tabular}
\end{table*}


Classical optimal transport models require that every piece of mass in the source distribution is transported to an equal-weight piece of mass in the target distribution. However, this requirement is too restrictive for varying-length document classification, especially when there are words semantically far away from the motifs of the documents.
The following example illustrates that the semantic outliers correspond to distant points  and can mislead the output of WMD. 

\begin{example}~\label{Example:WMD}
	Consider the three sentences in Figure~\ref{fig-demo}. Indeed, sentence A has positive semantics while B and C are negative. Therefore, well-defined document distance should reveal $D_{AB} > D_{BC}$.
	
	After removing stop words, the cost to transport from B to A or C is listed in Table \ref{Table:example}. 
	During the transport from B to C, the mass at ``awful'' in B is equally allocated to the five words in C. Since four out of the five words are semantically far from ``awful'', the average individual cost is pulled up, which makes  $\mathrm{WMD}(A, B) < \mathrm{WMD}(B,C)$.
	\vskip-2em
\end{example}

Example \ref{Example:WMD} shows that WMD tends to overestimate the semantic dissimilarity when the longer document contains additional details that not involved in the shorter one. In this length-varying case, WMD may not be an effective metric for comparing documents with rich semantic details. Especially, this situation becomes extremely severe in some advanced tasks such as text summarizing~\cite{DBLP:conf/ijcai/KedzieM16}, title generation~\cite{DBLP:conf/ijcai/ZhangSHNWDCY17} and length-varying matching~\cite{DBLP:conf/acl/BhatSGX18}.

Supervised Word Mover's Distance (S-WMD)~\cite{DBLP:conf/nips/HuangGKSSW16} tried to partly alleviate this overestimation issue by introducing global modification. 
S-WMD introduced the histogram importance vector to re-weight the nBOW distribution.
The re-weighted parameters in S-WMD do not rely on any specific texts but the training corpus.
However, as shown in Example~\ref{Example:WMD}, a reasonable re-weight mechanism should reduce the additional detail, which should be determined text-specifically.

To address the issues above, we introduce a robust document distance
based on the Wasserstein-Fisher-Rao (WFR) metric, a natural extension of Wasserstein metric newly developed from the theory of unbalanced optimal transport \cite{Kondratyev2016,DBLP:journals/siamma/LieroMS16,DBLP:journals/focm/ChizatPSV18,DBLP:journals/moc/ChizatPSV18}.
Unlike traditional Wasserstein metric, 
WFR metric allows transport from a piece of mass to another piece with different mass by adding a penalty term accounting for the unbalanced mass.
WFR document distance allows the unbalanced transport among semantic words, which naturally re-weight the transport plan based on the squared distances in word embedding space. This unique property of WFR alleviates the overestimation effects caused by WMD in a text-specific way. The following Example~\ref{Example:WFR} illustrates how WFR document distance remains effective in the case where WMD fails.
 
\begin{example}~\label{Example:WFR}
	The unbalanced transport plan from B to A or C and its cost that derives WFR document distance are listed in Table \ref{Table:example}.
	As we can see, the points closer to ``awful'', such as ``sad'', are more preferable in the transport plan from B to C.
	This effect naturally re-weights the five words in C and the distance of ``awful'' to them, making the total cost to transport from B to C lower than B to A.
	\vskip-1em
\end{example}

The main contributions of this paper are three folds.
\begin{itemize}
	\item The Wasserstein-Fisher-Rao metric is applied for measuring document distance. Theoretically, this new WFR document distance is highly interpretable but more effective than WMD and has only one hyper-parameter which is not sensitive.
	\item
	An effective pruning strategy is designed for fast top-k smallest WFR document distance query. Combined with GPU implementation, the computation efficiency is improved nearly by an order of magnitude (analysis can be found in the Appendix).
	\item We conduct extensive experiments in the tasks of varying-length matching and document classification.
	WFR document distance is proved to be far more robust than WMD when applied to varying-length documents.
	Moreover, the results of the eight document classification tasks comprehensively show the advantage of the WFR document distance. Finally, we show other frameworks based on metric space (for example WME) could be benefited from WFR document distance.
\end{itemize}

%% file: src/related.tex
\section{Related Work}
In this section, we briefly review the literature from the following three perspectives.

{\noindent \bf (a) Representation of documents.}
There have been many ways for documents representation. 
Latent Semantic Indexing~\cite{DBLP:journals/jasis/DeerwesterDLFH90} and Latent Dirichlet Allocation~\cite{DBLP:journals/jmlr/BleiNJ03} are based on inferred latent variables generated by the graphical model. However, most of those models are lack of the semantic information in the word embedding space~\cite{DBLP:conf/nips/MikolovSCCD13}. Stack denoising auto encoders~\cite{DBLP:conf/icml/GlorotBB11}, Doc2Vec~\cite{DBLP:conf/icml/LeM14} and skip-thoughts~\cite{DBLP:conf/nips/KirosZSZUTF15} are neural network based similarities. Despite their numerical success, those models are difficult to explain, and the performance always relies on the training samples. 

Recently, WMD~\cite{DBLP:conf/icml/KusnerSKW15} is proposed as an implicit document representation. By considering each document as a set of words in the word embedding space, it defines the minimal transportation cost as the distance between two documents. This metric is interpretable with the consideration of semantic movements. Many other metric learning models are inspired by the metric property of WMD.
S-WMD~\cite{DBLP:conf/nips/HuangGKSSW16} employed the derivative of WMD to optimize the parameterized transformation in word embedding space and histogram importance vector. Word Mover's Embedding~\cite{DBLP:conf/emnlp/WuYXXBCRW18} designed a kernel method on WMD metric space. However, those methods are still more or less suffer from the overestimation issue. They do not have the document-specific re-weight mechanism as WFR Document Distance.

{\noindent\bf (b) (Un)balanced optimal transport.}
Optimal transport (OT) has been one of the hottest topics of applied mathematics in the past few years. It is also closely related to some subjects in pure mathematics such as geometric analysis~\cite{MaTW2005,LottV2009} and non-linear partial differential equations~\cite{DBLP:journals/siamnum/FroeseO11, gu2013variational}. As the most fundamental and important object of OT, Wasserstein metric can be applied to measure the similarity of two probability distributions. The objective functions defined by this metric are usually convex, insensitive to noise, and can be effectively computed.
Thus, Wasserstein metric has been deeply exploited by many researchers and has been successfully applied to machine learning~\cite{DBLP:conf/icml/ArjovskyCB17}, image processing~\cite{DBLP:journals/corr/abs-1708-01955} and computer graphics~\cite{DBLP:journals/tog/SolomonGPCBNDG15}.

A key condition of Wasserstein metric is that the total mass of the measures to be compared should be identical. This requirement prevents further application of Wasserstein metric as it cannot capture the features with mass difference, growth or decay.
To overcome the shortage, WFR metric is proposed~\cite{Kondratyev2016,DBLP:journals/siamma/LieroMS16,DBLP:journals/focm/ChizatPSV18,DBLP:journals/moc/ChizatPSV18}
and applied to the situations where the similarity of objects (distributions) cannot be characterized by transport alone. 
Thus, it is not surprising that WFR has shown great performance in many applications, e.g. image processing~\cite{DBLP:journals/moc/ChizatPSV18} and tumor growth modeling~\cite{DiMarinoC2017}.

{\noindent \bf {(c) Fast calculation of (un)balanced optimal transport.}}
Sinkhorn algorithm~\cite{DBLP:conf/nips/Cuturi13} solves the entropy regularized OT problems. 
By reducing the entropy regularization term, the solution of each Sinkhorn iteration approximates to that of the original OT problem. 
A greedy coordinate descent version of Sinkhorn iteration~\cite{DBLP:conf/nips/AltschulerWR17} called Greenkhorn is proposed to improve the convergence property. 
Recently, Sinkhorn algorithm is applied to solve the unbalanced optimal transport problem~\cite{DBLP:journals/moc/ChizatPSV18} with modification on log-domain stabilization. In the case of document classification, an approximate solution of WFR is sufficient to serve as a good metric for documents.
Furthermore, as the dual problem of each sinkhorn iteration is computationally cheap and provides the lower bound of WFR Document Distance distance, we can further accelerate the KNN by introducing a pruning strategy.


%% file: src/method.tex
\section{Methods}

\subsection{Introduction of WFR metric}
Like traditional Wasserstein metric, WFR metric can be interpreted as the \emph{square root} of the minimum cost of a transport problem.
The most intuitive approach to formulate this optimization is by introducing the Benamou-Brenier formulation of optimal transport theory:

\begin{definition}[WFR metric]\label{def-WFR}
Given two measures $\mu$ and $\nu$ over some metric space $(X, \lVert \cdot \rVert)$ and $\eta>0$. Then the WFR metric is defined by the following optimization problem
\begin{align}
\mathrm{WFR}_{\eta}&(\mu,\nu) =
\bigg(\inf_{\rho,v,\alpha} \int_0^1 \int_\Omega \Big(\frac{1}{2} \|v(t,x)\|^2+ \frac{\eta^2}{2} \alpha(t,x)^2\Big) \rd x \rd t \bigg)^{\frac{1}{2}}
\end{align}
The infimum is taken over all the triplets $(\rho, v, \alpha)$ satisfying the following continuity equation:
\begin{align}
\partial_t \rho + \nabla \cdot(\rho v) = \rho \alpha, \quad
\rho(0,\cdot)=\mu, \quad \rho(1,\cdot) =\nu.
\end{align}

\end{definition}

The ``source term'' $\rho \alpha$ in the continuity equation and the corresponding penalty term
$\eta^2 \alpha(t,x)^2 / 2$ in the objective function in the formulation of WFR metric are the main differences between WFR and classical Wasserstein metric. They quantify the failure of conservation law (mass balance) in the transport plan. The parameter \(\eta\) controls the interpolation of the transport cost and the penalty term, which also determines the maximum distance that transport could occur. One can refer to \cite{DBLP:journals/focm/ChizatPSV18} for more details.

\subsubsection{Discrete WFR metric}
Discrete measure $ \mu $ over $ \mathbb{R}^n $ could be considered as $ \mu = \sum_i \mu_i \delta_{x_i} $, where $ \delta_{x} $ is the Dirac function on $ x \in \mathbb{R}^n $. When $ \sum_i \mu_i = 1 $, $ \mu $ is probabilistic distribution.
In the following context, we begin with the explicit formula of the transport between two Diracs and the proof is from Section 4 in \cite{DBLP:journals/focm/ChizatPSV18}.
\begin{lemma}\label{lemma-dirac2dirac}
	Given two Diracs of mass $h_0$ and $h_1$ and location $x_0$ and $x_1$, the WFR metric between them is
	\begin{align} \mathrm{WFR}_\eta(h_0 \delta_{x_0},h_1 \delta_{x_1}) = \nonumber \sqrt{2} \eta \Big[h_0 + h_1 - 2 \sqrt{h_0 h_1} \cos_+ \Big(\frac{|x_1 - x_0 |}{2 \eta} \Big) \Big]^{\frac{1}{2}},
	\end{align}
	where
	\begin{equation}
	\begin{aligned}
	\cos_+(x) =\begin{cases}
	\cos(x), &  x \in [-\pi/2,\pi/2];\\
	0 , & x \notin [-\pi/2,\pi/2].
	\end{cases}
	\end{aligned}
	\end{equation}
\end{lemma}
In general, the transport of two distributions composed of multiple Diracs can be interpreted as the linear combination of point-to-point transports. Considering two distributions,
\begin{align}
\mu = \sum_{i = 1}^I \mu_i \delta_{x_i}, \quad
\nu = \sum_{j = 1}^J \nu_j \delta_{y_j}, \quad \mu_i\ge0,\;\nu_j\ge0,
\end{align}
The mass $\mu_i,\nu_j$ are split into different pieces $\alpha_{ij} \geq 0, \beta_{ji} \geq 0$ as
\begin{equation}\label{wfr-from-dirac-constrain}
\sum_{j=1}^J \alpha_{ij} = \mu_i, \quad i = 1, \dots, I,
\quad \quad
\sum_{i=1}^I \beta_{ji} = \nu_j, \quad j = 1, \dots, J,
\end{equation}
and assign each pair of $(\alpha_{ij}, \beta_{ji})$ to the transport between $x_i$ and $y_j$. The WFR distance between $\mu$ and $\nu$ is 
\begin{equation}\label{wfr-from-dirac}
\begin{aligned}
\mathrm{WFR}_\eta^2(\mu,\nu) = & \min_{\alpha_{ij},\beta_{ji}}\sum_{i,j}  \mathrm{WFR}_\eta^2(\alpha_{ij} \delta_{x_i}, \beta_{ji} \delta_{y_j})\\
\mbox{ s.t.}\quad & \text{$\alpha_{ij}$ and $\beta_{ji}$ satisfy \eqref{wfr-from-dirac-constrain}.}
\end{aligned}
\end{equation}
%
%
%
%
It is noted that the problem of \eqref{wfr-from-dirac} is equivalent to the minimization problem in Definition~\ref{def-WFR}. However, it is difficult to find a numerical method to implement \eqref{wfr-from-dirac}. By taking dual form and changing variables alternatively, Theorem~\ref{thm-discrete-wfr} which is more numerically friendly is derived.
\begin{theorem}\cite{DBLP:journals/focm/ChizatPSV18} \label{thm-discrete-wfr}
Wasserstein-Fisher-Rao metric $ \mathrm{WFR}_\eta(\mu, \nu) $ for two discrete measures $\mu$, $\nu$ is the optimum of the primal problem:
\begin{align}\label{wfr-primal}
\mathrm{WFR}_{\eta}(\mu, \nu) = \inf_{R_{ij}\geq 0} J_{\eta}(R; \mu, \nu).
\end{align}
$ R_{ij} $ is the transport plan and the objective function $ J_{\eta} $ is 
\begin{align}\label{wfr-primal-obj}
J_{\eta}(R; \mu, \nu) =
\sum_{i,j} C_{ij} R_{ij} + \mathcal{KL}\left(\sum_j R_{ij} \| \mu\right) + \mathcal{KL}\left( \sum_i R_{ij} \| \nu\right)
\end{align}
where 
\begin{align}\label{costfunc}
C_{ij} = -2\log( \cos_+( |x_i - y_j| /2 \eta ))
\end{align}
is the cost matrix and $\mathcal{KL}$ denotes the KL divergence.\footnote{Applying this cost function in balanced OT is another modification. We did the ablation study in Section~\ref{matching} to show that the KL part is also necessary.}
The corresponding dual problem is
\begin{align}\label{wfr-dual}
\sup_{\phi_i,\psi_j} D_{\eta}(\phi, \psi; \mu, \nu)
\quad
s.t.& \quad \phi_i + \psi_j \leq C_{ij} \; \textnormal{ for any } \; i,j.
\end{align}
where the dual objection function is
\begin{equation}\label{wfr-dual-obj}
D_{\eta}(\phi, \psi; \mu, \nu) =\sum\limits_i \left(1 \!- \!e^{-\phi_i}\right) \!\mu_i + \sum\limits_j \!\left(1 \!- \!e^{-\psi_j}\right) \!\nu_j.
\end{equation}

\end{theorem}

\subsubsection{Sinkhorn iteration for WFR metric}
Sinkhorn iteration aims at solving the family of ``entropy regularized'' optimal transport problems. We use the calligraphy letter to distinguish the regularized problem from the original one.
The entropy regularized optimal transport problem is the minimization of 
\begin{align}\label{wfr-regularized}
\inf_{R_{ij}>0} \mathcal{J}_{\eta,\epsilon} (R):=J_{\eta}(R) + \epsilon \sum_{ij} R_{ij} \log(R_{ij}),
\end{align}
which is strictly convex.
Up to a multiplier $2 \eta^2$, we have
\begin{align}
\mathcal{J}_{\eta,\epsilon}(R) = \mathcal{KL}\Big(\sum_j R_{ij} \Big\| \mu\Big) + \mathcal{KL}\Big(\sum_i R_{ij} \Big\| \nu\Big) + \epsilon \mathcal{KL}\big(R_{ij} || \exp(-C/\epsilon)\big)
\end{align}
By convex optimization theory~\cite{Rockafellar1970Convex}, the dual problem of \eqref{wfr-regularized} is 
\begin{equation}\label{sinkhorn-dual}
\sup_{\phi, \psi}  \mathcal{D}_{\eta,\epsilon}(\phi, \psi),
\end{equation}
where $K_{\epsilon} = e^{-C/\epsilon}$, $(\phi \oplus \psi)_{ij} = \phi_i + \psi_j$ and
\begin{equation}
\mathcal{D}_{\eta,\epsilon}(\phi, \psi)=\langle 1-e^{-\phi}, \mu \rangle + \langle 1-e^{-\psi}, \nu \rangle 
+ \epsilon \langle 1 - e^{\frac{\phi \oplus \psi}{\epsilon}}, K_{\epsilon} \rangle.
\end{equation}
%
The WFR Sinkhorn iteration $ S_{\epsilon} $ solves problem~\eqref{sinkhorn-dual} for fixed $ \epsilon $.
The $ \phi $ and $ \psi $ are updated by Bregman iteration~\cite{DBLP:journals/siamsc/BenamouCCNP15} alternatively, i.e.
\begin{equation}
\begin{aligned}\label{eq:iter-prob}
& \phi^{(l+1)}=\argmax_{\phi} \langle 1\!-\!e^{-\phi},\mu\rangle+\epsilon\langle 1\!-\!e^{\frac{\phi\oplus \psi^{(l)}}{\epsilon}},K_{\epsilon}\rangle, \\ 
& \psi^{(l+1)}=\argmax_{\psi}\langle 1-e^{-\psi},\nu \rangle + \epsilon \langle 1 - e^{\frac{\phi^{(l+1)} \oplus \psi}{\epsilon}},K_{\epsilon} \rangle.
\end{aligned}
\end{equation}
Those two subproblems could be solved in Proposition~\ref{prop:iteration-step}.
\begin{proposition}\label{prop:iteration-step}
	Let $u = e^{\phi/\epsilon}$ and $v = e^{\psi/\epsilon}$, the analytical solution of subproblems in Equation~\eqref{eq:iter-prob} is
	\begin{equation}\label{iter-step}
	\begin{aligned}
	u^{(l+1)}_i &= \left( \mu_i / \textstyle \sum_j e^{-C_{ij}/\epsilon} v^{(l)}_j \right)^{1/(1+\epsilon)},
	v^{(l+1)}_j &= \left( \nu_j / \textstyle \sum_i e^{-C_{ij}/\epsilon} u^{(l+1)}_i \right)^{1/(1+\epsilon)}.
	\end{aligned}
	\end{equation}
	where $ i = 1, \dots, I $ and $ j = 1, \dots, J $.
	Equation~\eqref{iter-step} is the iteration step solves Equation~\eqref{wfr-regularized}.
\end{proposition}

The details of the Sinkhorn algorithm for WFR distance is given in Appendix.
It is noted that in \eqref{iter-step}, the term  $ e^{-C_{ij}/\epsilon} $ or $ u, v $ might be extremely small or large which could cause the numerical instability in the implementation. In the Sinkhorn algorithm, $ \exp((\phi_i + \psi_j - C_{ij})/\epsilon) $ is taken as a whole for improving the numerical stability.
To solve the original problem~\eqref{wfr-primal}, we sequentially perform WFR Sinkhorn iteration $ \{S_{\epsilon_n}\} $ on descending $ \{\epsilon_n\} $ where $ \epsilon_n \to 0 $, and adopt the optimal $ \phi, \psi $ for $ S_{\epsilon_n} $ as the initial value for $ S_{\epsilon_{n+1}} $. The precision of WFR metric is controlled by the gap between the primal and dual problem.

\subsection{WFR Document Distance}
\subsubsection{Approximate WFR document distance}\label{sec:approx-wfr-doc-dist}

To apply WFR document distance, one document should be formulated as one discrete measure $ \mu = \sum_{k = 1}^K \mu_k \delta_{x_k} $.
Following the bag of words representation, a document $ D $ is considered as a multi-set with $ K $ elements $ D = \{w_1, \dots, w_K\} $ and the number of occurrence of each word $ C_D = \{c_1, \dots, c_K\} $.
Each word $ w_i $ belongs to the vocabulary $ \mathcal{V} $.
The nBOW distribution is defined by normalizing the number of occurrences: $ \mu_k = c_k / \sum_j c_j $ for $ k = 1, \dots, K $.
Given a word embedding $ \mathcal{X}: \mathcal{V} \mapsto \mathbb{R}^n $, each word $ w_k $ in Document $ D $ is mapped to a point in $ \mathbb{R}^n $, i.e.\ $ x_k = \mathcal{X}(w_k) $ for $ k = 1, \dots, K $. Formally, we define the WFR document distance as follows.
\begin{definition}[WFR document distance]\label{wfr-document-distance}
	Given a pair of documents $D_1$ and $D_2$ and a constant $\eta>0$. Let $ \mu = \sum_{i=1}^I \mu_i \delta_{x_i} $ and $ \nu = \sum_{j=1}^J \nu_j \delta_{y_j} $ be the nBOW probability distribution of $D_1$ and $D_2$ respectively. The WFR document distance between $D_1$ and $D_2$ is defined as
	\begin{equation}
	\mathrm{Dist}(D_1,D_2) = \mathrm{WFR}_\eta(\mu,\nu).
	\end{equation}
\end{definition}
The numerical method for calculating the WFR document distance is present in Appendix. In our experiment, we use $ M=5 $ WFR Sinkhorn iterations with parameter $\{(\epsilon_m, n_m) = \{(e^{-m-1}, 32m)\}\}$ for the $ m $-th iteration.
Experiments show that the mean relative error of the approximate solution is no more than $ 0.001 $ by evaluating the duality gap which achieves the desired accuracy. 

\subsubsection{Pruning strategy for top-k smallest WFR document distance query}
Top-k smallest WFR document distance query is significant in applications like document retrieval.
\cite{DBLP:conf/icml/KusnerSKW15} proposed a pruning strategy for fast WMD-KNN classification based on the lower bound of WMD.
In the case of WFR document distance, it is natural to adopt the evaluated value of the dual objective function~\eqref{wfr-dual-obj} as a lower bound.
With the descending of the entropy regularization's coefficient $ \epsilon $, the dual lower bound gets more and more tight. 

In the  top-k smallest WFR document distance query setting, the query document $ D_0 $ is formulized as $ \mu_{D_0} = \sum_{i=1}^I \mu_i \delta_{x_i} $ and the document samples are $ \{(D_n, y_n)\}_{n=1}^N $ where each $ D_n $ as  $ \nu_{D_n} = \sum_{j=1}^J \nu_j^{(n)} \delta_{y_j^{(n)}} $, $ n=1, \dots, N $.
Considering the task with hyper-parameter $ k $, after each WFR Sinkhorn iteration, we sort the document samples by the value of primal objective~\eqref{wfr-primal} and take the maximum of WFR document distance among the first $ k $ smallest values as the threshold.
Furthermore, we evaluate the dual lower bound, document samples with lower bounds that are larger than the threshold will be dropped.
By this way, we only need to perform few WFR Sinkhorn iterations for most of the samples, which saves a lot of time.

For WFR document distance described in Definition ~\ref{wfr-document-distance}, the number of WFR Sinkhorn iterations $ M $ and parameters $ \{(\epsilon_m, n_m)\} $ for each WFR Sinkhorn iteration is fixed.
Given document size $ L $, the time complexity of the Sinkhorn iteration is $ O(L^2) $ for a fixed parameter.
Given the size of training samples $ N $, the time complexity of WFR-KNN classification is bounded by $ O(NL^2) $.
It is noticed that this asymptotic bound cannot be further improved since the time complexity of the distance/cost matrices calculation between the evaluated sample and $ N $ labeled samples are $ O(NL^2) $.
In Appendix we demonstrate the details of top-k smallest WFR document distance query with pruning strategy.

%% file: src/discuss.tex
\section{Experiment and Discussion}



In this section, we demonstrate the supreme of WFR Document Distance over WMD and other WMD based metrics in two tasks.
The first task directly illustrates the robustness of WFR over WMD when matching length-varying documents.
The second task examines the effectiveness of WFR Document Distance on a vast number of documents by KNN classification.
The WMD is computed by the code provided by~\cite{DBLP:conf/icml/KusnerSKW15}.

\subsection*{Task 1: Length-varying matching}\label{matching}

\textbf{(a) Setup}. The concept-project dataset by~\cite{DBLP:conf/acl/BhatSGX18} is designed for length-varying document matching task.
This dataset contains 537 samples.
Each sample contains one short document named ``concept'', one long document named ``project'' and one human annotated binary label for whether this pair is a good match.
The length of each ``concept'' and corresponding ``project'' varies a lot. The mean of the distinct words among all ``concept'' is 26.4, while the mean distinct words among all ``project'' is 556.6.
The matching is binary classification. The ratio of true and false label is 56:44.
For this task, we take WMD as the baseline.
It has been proven~\cite{DBLP:conf/acl/BhatSGX18} that WMD is a stronger than the neural network methods such as doc2vec~\cite{DBLP:conf/icml/LeM14}. We also apply the cost function of WFR (see Equation~\eqref{costfunc}) to balanced optimal transport as an ablation study. 
Suggested by~\cite{DBLP:conf/acl/BhatSGX18}, the document distance between ``concept'' and ``project'' (WMD or WFR) is used as one score of the concept-project pair for binary classification.
Given the threshold, the pair with the distance smaller than the threshold is classified as the true label.
The word embedding in the experiments is pretrained by fasttext~\cite{mikolov2018advances}.
We evaluate WMD and WFR Document Distance on the whole dataset.
After calculating the document distance of each pair, we adjust the threshold to obtain the precision-recall curve.

\begin{figure}[t]
	\centering
	\includegraphics[width=0.6\columnwidth, height=6cm]{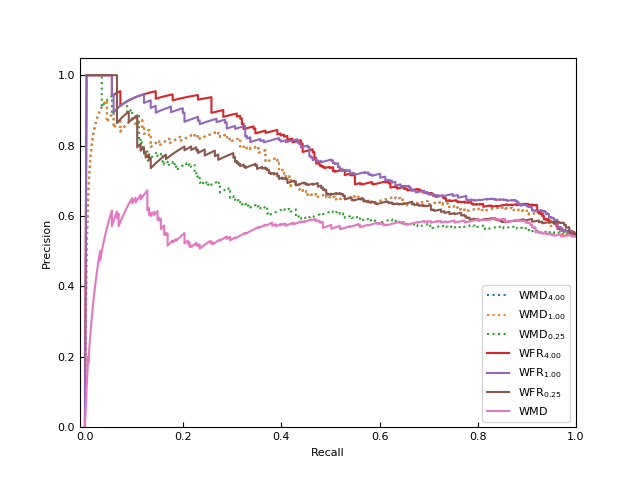}
	\vskip-1em
	\caption{PR curve for the length-varying matching task}
	\label{fig:length-varying-match}
	\vskip-1em
\end{figure}

\textbf{(b) Discussion}.  Figure~\ref{fig:length-varying-match} illustrates the precision-recall curves of WMD and WFR Document Distance whose hyper-parameter $ \eta $ ranges from $0.25$ to $4$.
The curve of WMD is dominated by that of all WFR Document Distances at all recall level.
At low recall level (less than 0.1, the threshold is small), the pairs with small document distances are classified to be good matches. The high precision (over 0.8) of WFR Document Distances of all hyper-parameters shows the effectiveness of our WFR Document Distance. The low precision of WMD is consistent with the observation in the Example~\ref{Example:WMD} that document pair who is semantically similar may not be closed under WMD.
WFR Document Distance is proved to be more reliable and robust than WMD and is not sensitive to the hyper-parameter $ \eta $. ($ \mathrm{WMD}_{4.00} $ and $ \mathrm{WMD}_{1.00} $ collapsed together, which also supports $ \eta $ is not sensitive.)
Composing ground metric of WMD with the cost function of WFR improves the performance. However, for fixed parameter $ \eta $ this amendment in balanced optimal transport is clearly weaker than unbalanced WFR document distance.

\subsection*{Task 2: KNN classification}

\begin{table*}
	\centering
	\caption{KNN classification error rate for WFR and other baselines and combine with WME(512).}\vskip-.5em
	\label{table:knn}
	{\scriptsize \begin{tabular}{p{1.8cm}p{1cm}p{1.2cm}p{1.2cm}p{1cm}p{1cm}p{1cm}p{1cm}p{1cm}}
		\toprule
		DATASET       & BBCSPORT             & TWITTER               & RECIPE                & OHSUMED         & CLASSIC              & REUTERS        & AMAZON               & 20NEWS          \\ \hline
		WMD           & $4.6\pm0.7$          & $28.7\pm0.6$          & $42.6\pm0.3$          & $44.5$          & $2.8\pm0.1$          & $3.5$          & $7.4\pm0.3$          & $28.3$          \\
		S-WMD         & $2.1\pm0.5$          & $27.5\pm0.5$          & $39.2\pm0.3$          & $\mathbf{34.3}$ & $3.2\pm0.2$          & \textbf{$3.2$} & $5.8\pm0.1$          & $26.8$          \\
		WFR           & $\mathbf{0.8\pm0.3}$ & $\mathbf{26.4\pm0.2}$ & $\mathbf{38.9\pm0.1}$ & $41.82$         & $\mathbf{2.6\pm0.2}$ & $\mathbf{3.2}$ & $\mathbf{4.8\pm0.2}$ & $\mathbf{22.3}$ \\ \hline
		WME(512)+WMD  & $3.5\pm0.7$          & $26.8\pm2.3$          & $48.0\pm0.6$          & $42.1$          & $4.8\pm0.3$          & $4.0$          & $7.4\pm0.4$          & $ 30.7 $        \\
		WME(512)+WFR  & $2.7\pm1.0$          & $26.0\pm1.9$          & $43.3\pm1.2$          & $37.2$          & $3.7\pm0.4$          & $3.7$          & $7.5\pm0.3$          & $ 29.8 $        \\ \bottomrule
	\end{tabular}}
\vskip-2em
\end{table*}

\textbf{(a) Setup}. We evaluate the effectiveness of WFR Document Distance on eight document classification datasets: BBCSPORTS: BBC sports article at 2004-2005; TWITTER: sentiment classification corpus of tweets; RECIPE: recipe procedures from different origins; OHSUMED: medical abstracts from cardiovascular disease groups; CLASSIC: academical papers by different publishers; REUSTERS and 20NEWS: news articles by topics. The preprocessing procedures and the choice of word embeddings are the same as that described by~\cite{DBLP:conf/icml/KusnerSKW15, DBLP:conf/nips/HuangGKSSW16}. We use directly the preprocessed version of datasets from the authors.
The key information of the datasets are presented in Appendix, including the number of train/test samples and the average and the standard deviation of the number of distinct words (NDW).
Besides WMD, We consider an additional supervised baseline named
\textbf{Supervised Word Mover's Distance (S-WMD)}. Compared to WMD, this method employed a histogram importance vector $ w $ of vocabulary to re-weight the nBOW distribution $ \tilde{f}_i = w_i f_i /\sum_j w_jf_j $, and a linear transformation $ A: x_i \mapsto Ax_i $ to modify the distances in the word embedding space. The parameters are trained by gradient descent of the loss defined by Neighborhood Components Analysis (NCA).
Other traditional document representation or similarity baselines are proved to be significantly weaker than WMD and SWMD~\cite{DBLP:conf/icml/KusnerSKW15,DBLP:conf/nips/HuangGKSSW16}. So they are not included.
Throughout the experiments, we optimize over the neighborhood size $(k \in \{1, \dots,19\})$ in KNN and the only hyper-parameter ($\eta \in \{1, 1/2, 1/3, 1/4\}$) by 5-fold cross-validation. We obtain the original code from the authors and re-conduct the evaluation process. For datasets without predefined train/test splits  (bbcsport, twitter, recipe, classic, amazon), we report the mean and standard deviation of the performance over five random 70/30 train/test splits.

\textbf{(b) Discussion}. In the first three rows of Table~\ref{table:knn} we output the results from three different document distances and eight datasets.
Firstly, we compare the performance between WFR with WMD.
As presented, WFR Document Distance has less KNN classification error rate at all datasets. Furthermore, for the datasets with large standard deviation of NDW (exceeds 40, see Appendix), i.e. dataset BBCSPORTS, AMAZON and 20NEWS, WFR outperforms the document distance with a clear margin.
For those datasets with less standard deviations of NDW, the reduction of the KNN classification error is not that significant.
Secondly, we compare the performance between WFR with S-WMD. 
WFR successfully outperforms S-WMD in six out of eight datasets even though S-WMD has more supervised parameters.
The successful of WFR over S-WMD since a more effective way to re-weight the transport plan is automatically captured by WFR, rather than text-independent global re-weighting in S-WMD.
We notice that S-WMD only outperforms WFR and WMD at OHSUMED dataset.
The medical term for cardiovascular disease in the OHSUMED dataset may not have proper word vector.
The text-independent deficiency of the word embedding might be relieved by supervision in S-WMD.

\textbf{(c) WFR Document Distance for Other Frameworks}. Word Mover's Embedding (WME~\cite{DBLP:conf/emnlp/WuYXXBCRW18}) framework is proposed to abstract the document space of Word Mover's Distance (or other metric spaces). This framework realized fast estimation of WMD by Monte Carlo's method. We found that replacing the WMD in WME framework with WFR document distances effectively improves the original results. Last two rows of Table~\ref{table:knn} compare the effect of WFR and WMD in WME framework with 512 samples. WME+WFR consistently outperforms WME+WMD under exactly the same setting (512 MC samples). Notably, the results of WME+WFR are closed to those of WME+WMD reported with 8 times MC samples (4096) with minor computation cost (see Appendix).

%% file: src/conclusion.tex
\section{Conclusion}

In this paper, the Wasserstein-Fisher-Rao metric is applied as one unsupervised document distance (WFR document distance) which is demonstrated to be theoretically solid, easy to interpret and proved to be much more robust than WMD.
WFR and its derivatives could be calculated efficiently by WFR Sinkhorn iterations with GPU acceleration.
Similar to its ancestors, WFR document distance benefits from the semantic similarity of word embedding space while employs automatically re-weighted transport plan overcome the overestimation issue appearing in varying-length situations.
Numerical expriments confirm the effectiveness and efficiency of the new proposed metric.

%% file: src/appendix.tex
\newpage
\appendix
\section{Algorithms}

Algorithm~\ref{alg:log-domain-stabilized-sinkhorn} describes single Sinkhorn iteration that are used to calculate entropy regularized Wasserstein-Fisher-Rao metric with log-domain stabilization. Algorithm~\ref{alg:wfr-document-distance} shows how to get the WFR document distance based on Algorothm~\ref{alg:log-domain-stabilized-sinkhorn}

\begin{algorithm}[h]
	\caption{WFRSinkhorn($ \mu, \nu, C, \epsilon, n, \phi, \psi $)}
	\label{alg:log-domain-stabilized-sinkhorn}
	\begin{algorithmic}
		\STATE {\bfseries Input:} \\
		\hspace{0.05\linewidth} Discrete measure $\mu$ and $\nu$, \\
		\hspace{0.05\linewidth} cost matrix $C$, \\
		\hspace{0.05\linewidth} $ \epsilon $ for entropy regularization and number of iteration $ n $, \\
		\hspace{0.05\linewidth} dual potential $ \phi $ and $ \psi $\\
		\STATE {\bfseries Output:}\\ 
		\hspace{0.05\linewidth} Optimal transport plan $ R $ and potential $ \phi $, $ \psi $.
		\IF{$ \phi $ is None or $ \psi $ is None}
		\STATE $ (b, \phi, \psi) \gets (\mathbf{1}_J, \mathbf{0}_I, \mathbf{0}_J) $
		\ELSE
		\STATE $ (b, \phi, \psi) \gets (\mathbf{1}_J, \phi, \psi) $
		\ENDIF
		\STATE $ R_{ij} \gets \exp(\frac{\phi_i + \psi_j - C_{ij}}{\epsilon}) $ 
		\FOR{$k=1$ {\bfseries to} $n$}
		\STATE $ a_i \gets ( \mu_i / \exp(\phi_i) \sum_j R_{ij} b_j ) ^ {1/(1+\epsilon)} $
		\STATE $ b_j \gets ( \nu_j / \exp(\psi_j) \sum_i R_{ij} a_i ) ^ {1/(1+\epsilon)} $
		\IF{$ \lVert a \rVert $ or $ \lVert b \rVert $ is too large, or $ k $ equals to $ n $}
		\STATE $ \phi \gets \phi + \epsilon \log(a) $
		\STATE $ \psi \gets \psi + \epsilon \log(b) $
		\STATE $ R_{ij} \gets \exp(\frac{\phi_i + \psi_j - C_{ij}}{\epsilon}) $
		\STATE $ b \gets \mathbf{1}_J $
		\ENDIF
		\ENDFOR
		\STATE Return $ (R, \phi, \psi) $.
	\end{algorithmic}
\end{algorithm}

\begin{algorithm}[h]
	\caption{WFRDocDist($ \mu, \nu, M, \{(\epsilon_m, n_m)\}, \eta $)}
	\label{alg:wfr-document-distance}
	\begin{algorithmic}
		\STATE {\bfseries Input:} \\
		\hspace{0.05\linewidth} Documents distribution $ \mu $ and $ \nu $, \\
		\hspace{0.05\linewidth} number of the WFR Sinkhorn iteration $ M $, \\
		\hspace{0.05\linewidth} $ \{(\epsilon_m, n_m)\}_{m=1}^M $ for each iteration,  $ \eta $ for WFR metric.
		\STATE {\bfseries Output:}\\
		\hspace{0.05\linewidth} WFR document distance
		\STATE $ C_{ij} \gets -2\log\left(\cos_+\left(\frac{\lVert x_i - y^{(n)}_j \rVert_2}{2\eta}\right)\right) $
		\STATE $ (\phi, \psi) \gets (\text{None}, \text{None}) $
		\FOR{m from 1 to $ M $}
		\STATE $ (R, u, v) \gets \text{WFRSinkhorn}(\mu_{D_1}, \nu_{D_2}, C, \epsilon_m, n_m, \phi, \psi) $
		\ENDFOR
		\STATE Return $ J_{\eta}(R; \mu, \nu) $.
	\end{algorithmic}
\end{algorithm}

Algorithm~\ref{alg:wfr-knn} describe how to accelerate the KNN calculation by pruning the lower bounds.

\begin{algorithm}[t]
	\caption{Top-k smallest WFR document distance query}
	\label{alg:wfr-knn}
	\begin{algorithmic}
		\STATE {\bfseries Input:}\\ 
		\hspace{0.05\linewidth} Test document $ D_0 $ and training document set $ \{(D_n, y_n)_{n=1}^N\} $, \\
		\hspace{0.05\linewidth} number of iteration $ M $, parameter $ \{(\epsilon_m, n_m)\}_{m=1}^M $ for each WFR Sinkhorn iteration, \\
		\hspace{0.05\linewidth} $ \eta $ for WFR document distance and $ K $ for KNN.
		\STATE {\bfseries Output:}\\ 
		\hspace{0.05\linewidth} k indices of top-k smallest WFR document distance samples
		\FOR{each $ D_n $ in training set}
		\STATE $ C^{(n)}_{ij} \gets -2\log\left(\frac{\cos_+\left(\lVert x_i - y^{(n)}_j \rVert_2\right)}{2\eta}\right) $
		\STATE $ (u^{(n)}, v^{(n)}) \gets (\text{None},\text{None}) $
		\ENDFOR
		\STATE $ FilteredIndex \gets [1, \dots, N] $
		\FOR{$ m $ from 1 to $ M $}
		\STATE $ CandidateIndex \gets FilteredIndex $
		\STATE $ FilteredIndex \gets [\quad]$
		\STATE $ threshold \gets 0$ 
		\FOR{$ k $ from 1 to $ K $}
		\STATE $ t \gets $ \text{WFRDocDist}($ \mu_{D_0}, \nu_{D_m}, M, \{(\epsilon_m, n_m)\}, \eta $)
		\IF{$ t \ge threshold $}
		\STATE  $ threshold \gets t $
		\ENDIF
		\ENDFOR
		\FOR{ each $ i \in CandidateIndex  $}
		\STATE $ (R^{(i)}, u^{(i)}, v^{(i)}) \gets \text{WFRSinkhorn}(\mu_{D_0}, \nu_{D_i}$, $ C^{(i)}, \epsilon_m, n_m, u^{(i)}, v^{(i)}) $
		\IF{$ D_{\eta}(u^{(i)}, v^{(i)}; \mu_{D_0}, \nu_{D_i}) < threshold $}
		\STATE append $ i $ to $ FilteredIndex $
		\ENDIF
		\ENDFOR
		\STATE Sort $ FilteredIndex $ by $ J_{\eta}(R^{(i)}; \mu_{D_0}, \nu_{D_i}) $ in ascending order.
		\ENDFOR
		\STATE Return the first-$ K $ elements of $ FilteredIndex $.
	\end{algorithmic}
\end{algorithm}

\section{Information of Datasets}

The information of datasets for KNN evaluation is shown in Table~\ref{table:dataset_info} 

\begin{table}[h]
	\caption{The datasets used for evaluation and their description.}
	\label{table:dataset_info}
	\begin{center}
		\begin{sc}
			\begin{tabular}{crrrr}
				\toprule
				DATASET & \# TRAIN & \# TEST & AVG NDW & STD NDW \\
				\hline
				bbcsports&   517&  220& 117.0& \textbf{55.0} \\
				twitter  &  2175&  933&   9.9&  5.1 \\
				recipe2  &  3059& 1311&  48.4& 29.8 \\
				ohsumed  &  3999& 5153&  59.2& 22.3 \\
				classic  &  4965& 2128&  38.8& 27.7 \\
				reuters  &  5485& 2189&  37.1& 36.6 \\
				amazon   &  5600& 2400&  45.1& \textbf{45.8} \\
				20news   & 11293& 7528&  69.7& \textbf{70.1} \\
				\bottomrule
			\end{tabular}
		\end{sc}
	\end{center}
\end{table}

\section{Pruning Efficiency, GPU Acceleration and Time Cost}
The pruning strategy for top-k smallest WFR document distance query and GPU parallelism is important to constrain the computation cost of KNN in an affordable range.
Table~\ref{table:prune-efficiency} demonstrates the effect of prune strategy and GPU parallelism.

The columns in Table~\ref{table:prune-efficiency} named by Prune shows the average percent of samples left after $ m $-th round of prune.
For BBCSPORTS dataset, since the training set has only 517 samples, 3.87\% of the training set contains 20 samples, which is the minimal number required for KNN classifier when $ K = 20 $.
For other larger datasets, we noticed that after 2 rounds, more than 98\% of the training samples are pruned.
For all datasets, one could examine that after 3 rounds, the number of left samples is about 20, which is suitable for the following KNN classification.
With this pruning strategy, most of the computing cost is at the 1st Sinkhorn iteration, which is of time complexity $ O(NL^2) $.
In other words, one could improve the final precision for top-k smallest WFR document distance with merely little cost.

\begin{table}
	\caption{KNN prune efficiency and GPU acceleration ratio for eight datasets}

	\label{table:prune-efficiency}
	\begin{center}
		\begin{sc}
			\begin{tabular}{crrrr}
				\toprule
				DATASET   & \multicolumn{3}{c}{Prune} & GPU Acc. \\
				\cline{2-4}
				& 1st & 2nd & 3rd & Ratio \\
				\hline
				BBCSPORTS & 89.6\% & 6.2\% & 4.1\% & 3.9  \\
				TWITTER   & 2.1\%  & 1.0\% & 0.9\% & 35.0  \\
				RECIPE2   & 46.3\% & 1.9\% & 0.8\% & 7.3  \\
				OHSUMED   & 31.1\% & 0.8\% & 0.5\% & 8.9  \\
				CLASSIC   & 33.8\% & 1.2\% & 0.5\% & 9.0  \\
				REUTERS   & 3.2\%  & 0.6\% & 0.4\% & 11.8  \\
				AMAZON    & 1.7\%  & 0.5\% & 0.4\% & 9.9  \\
				20NEWS    & 40.0\% & 0.6\% & 0.2\% & 6.7  \\
				\bottomrule
			\end{tabular}
		\end{sc}
	\end{center}
\vskip-1em
\end{table}

\begin{figure}[t]
	\includegraphics[width=\columnwidth]{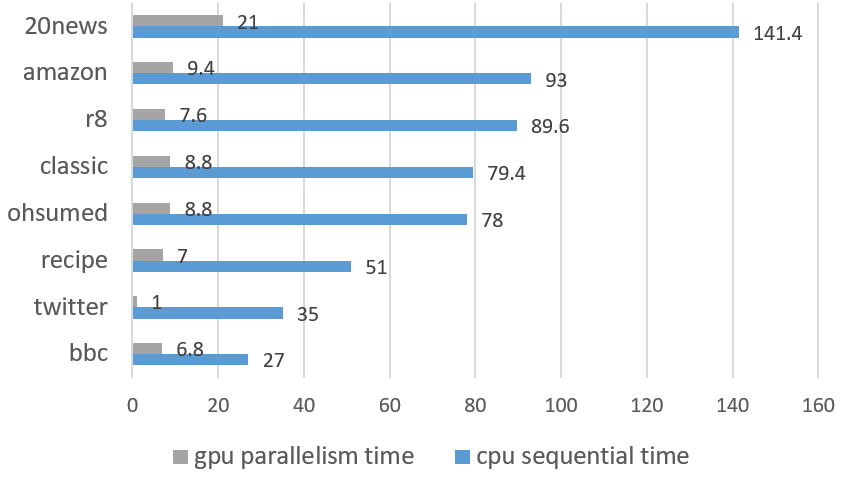}
	\vskip-1em
	\caption{Scaled time cost for one KNN classification (K=20)}
	\label{fig:cpu-gpu}
	\vskip-1em
\end{figure}

Figure~\ref{fig:cpu-gpu} shows the averaging time of one KNN classification on eight datasets. The value is scaled by the minimal time cost (TWITTER dataset by GPU).
The column in Table~\ref{table:prune-efficiency} named by GPU Acc. Ratio denotes the acceleration ratio. For example, 3.9 for bbcsports means that one CPU (Core i7-7700HQ) computation costs 3.9 times as one GPU (GTX-1080Ti).
For TWITTER, this dataset is too small so that all the data could be placed into the visual memory of GPU at single batch, which allows extremely high parallelism.
Discard the highest and lowest value of the acceleration ratio, we observe that the GPU parallelism provides about 8.9 times acceleration.

Another concerning about the computation time is the difference between WFR and WMD by Sinkhorn iteration. We evaluate conduct 1000 pairs of documents (point clouds) with number of distinct words varying from 10 to 1000. Experiments are conducted by MATLAB with fixed iteration parameters. WFR takes 42.3 seconds in total while WMD takes 39.7 seconds. We think about additional 5\% time cost is worthwhile.

\section{Detailed Evaluation of WFR and Other Framework}

Here we demonstrate the detailed results for WFR+WME and WMD+WME (WFR document distance and word mover's distance within WME framework). We present WME with two sizes of Monte Carlo samples, i.e. 512 and 4096. 
We didn't exact recover the original results in \cite{DBLP:conf/emnlp/WuYXXBCRW18} since we didn't know exact value of hyperparameters $ D_{max} $, $ \gamma $ and $ R $ that are used for each dataset. By similar parameter selection process, we produce the compatible results (WME(512)+WMD is close to WME(SR) in \cite{DBLP:conf/emnlp/WuYXXBCRW18} and WME(4096)+WMD is close to WME(LR)).
For WME+WFR, we take an additional cross-validation process to select the hyperparameter $ \eta $ of WFR.
The differences of WME and WFR are compared under the same MC sample condition. In most case, We could see that WME+WMD is weaker than WME+WFR for both 512 and 4096 MC samples. For some datasets (TWITTER, OHSUMED and CLASSIC), WME(512)+WFR is really close to the WME(4096)+WMD. This results support that WFR document distance is better than word mover's distance.

\begin{table}
	\centering
	\caption{KNN classification error rate for WME+WFR and WME+WFR.}
	\label{table:wfrwmewmd}
	{\scriptsize \begin{tabular}{p{1.8cm}p{1cm}p{1.2cm}p{1.2cm}p{1cm}p{1cm}p{1cm}p{1cm}p{1cm}}
			\toprule
			DATASET       & BBCSPORT             & TWITTER               & RECIPE                & OHSUMED         & CLASSIC              & REUTERS        & AMAZON               & 20NEWS          \\
 \hline
			WME(512)+WMD  & $3.5\pm0.7$          & $26.8\pm2.3$          & $48.0\pm0.6$          & $42.1$          & $4.8\pm0.3$          & $4.0$          & $7.4\pm0.4$          & $ 30.7 $        \\
			WME(512)+WFR  & $2.7\pm1.0$          & $26.0\pm1.9$          & $43.3\pm1.2$          & $37.2$          & $3.7\pm0.4$          & $3.7$          & $7.5\pm0.3$          & $ 29.8 $        \\ 
\hline				
			WME(4096)+WMD  & $2.0\pm1.0$          & $25.9\pm2.3$          & $40.2\pm0.6$          & $36.5$          &$3.0\pm0.3$          & $2.8$          & $5.7\pm0.4$          & $ 22.1 $        \\
			WME(4096)+WFR  & $1.8\pm1.0$          & $25.3\pm1.6$          & $40.1\pm0.6$          & $34.4$          & $2.9\pm0.3$          & $2.3$          & $5.5\pm0.5$          & $ 21.5 $        \\ 
			\bottomrule
	\end{tabular}}
	\vskip-2em
\end{table}